\documentclass[conference]{IEEEtran}
\IEEEoverridecommandlockouts
\usepackage{cite}
\usepackage{amsmath,amssymb,amsfonts}
\usepackage{algorithmic}
\usepackage{graphicx}
\usepackage{textcomp}
\usepackage{xcolor}

\usepackage{algorithm}
\usepackage{multirow}
\usepackage{amsthm}
\usepackage{booktabs}
\usepackage{subfigure}
\usepackage{enumitem}
\usepackage{tcolorbox}

\def\BibTeX{{\rm B\kern-.05em{\sc i\kern-.025em b}\kern-.08em
    T\kern-.1667em\lower.7ex\hbox{E}\kern-.125emX}}
\begin{document}

\title{Towards Efficient Resume Understanding: \\ A Multi-Granularity Multi-Modal Pre-Training Approach
\thanks{
* Corresponding author.\\
This work was accomplished when Feihu Jiang was an intern in Career Science Lab, BOSS Zhipin, supervised by Chuan Qin and Hengshu Zhu.\\

This work was supported in part by Guangzhou-HKUST Joint Funding Program(Grant No.2023A03J0008), Education Bureau of Guangzhou Municipality, Guangdong Science and Technology Department, Foshan HKUST Projects(FSUST21-FYTRI01A), Postdoctoral Fellowship Program of CPSF under Grant Number GZC20232811.
}
}

\author{
\IEEEauthorblockN{Feihu Jiang$^{1}$,\quad Chuan Qin$^{2,3*}$,\quad Jingshuai Zhang$^{2}$,\quad Kaichun Yao$^{4}$,\\
Xi Chen$^{1}$,\quad Dazhong Shen$^{5}$,\quad Chen Zhu$^{2}$,\quad Hengshu Zhu$^{2}$,\quad Hui Xiong$^{6,7,8*}$}
\IEEEauthorblockA{
\textit{$^1$University of Science and Technology of China,}\\
\textit{$^2$Career Science Lab, BOSS Zhipin,}\\
\textit{$^3$PBC School of Finance, Tsinghua University,}\\
\textit{$^4$Institute of Software, Chinese Academy of Sciences,}\\
\textit{$^5$Shanghai Artificial Intelligence Laboratory,}\\
\textit{$^6$The Thrust of Artificial Intelligence, The Hong Kong University of Science and Technology (GuangZhou),}\\
\textit{$^7$The Department of Computer Science and Engineering, The Hong Kong University of Science and Technology,}\\
\textit{$^8$Guangzhou HKUST Fok Ying Tung Research Institute,}\\
$\{$jiangfeihu, chenxi0401$\}$@mail.ustc.edu.cn, yaokaichun@outlook.com, xionghui@ust.hk\\
$\{$chuanqin0426, zhangjingshuai0, dazh.shen, zc3930155, zhuhengshu$\}$@gmail.com.
}
}


\maketitle

\begin{abstract}
In the contemporary era of widespread online recruitment, resume understanding has been widely acknowledged as a fundamental and crucial task, which aims to extract structured information from resume documents automatically. Compared to the traditional rule-based approaches, the utilization of recently proposed pre-trained document understanding models can greatly enhance the effectiveness of resume understanding. The present approaches have, however, disregarded the hierarchical relations within the structured information presented in resumes, and have difficulty parsing resumes in an efficient manner. To this end, in this paper, we propose a novel model, namely ERU, to achieve \textbf{e}fficient \textbf{r}esume \textbf{u}nderstanding. Specifically, we first introduce a layout-aware multi-modal fusion transformer for encoding the segments in the resume with integrated textual, visual, and layout information. Then, we design three self-supervised tasks to pre-train this module via a large number of unlabeled resumes. Next, we fine-tune the model with a multi-granularity sequence labeling task to extract structured information from resumes. Finally, extensive experiments on a real-world dataset clearly demonstrate the effectiveness of ERU.
\end{abstract}


\section{Introduction}
With the rapid growth and widespread adoption of online recruitment, employers receive over two hundred job applications for each available position~\cite{zety}. Recently, a multitude of intelligent recruitment techniques have emerged to expedite the selection of the most suitable candidate~\cite{rodney2019artificial}. Notably, the approach of resume understanding, also known as resume parsing, assumes a pivotal and indispensable role, as its ability to save recruiters from manually entering candidate information and instead automate the conversion of resumes into structured data. Such structured data can subsequently be utilized for various downstream applications, including person-job matching~\cite{qin2018enhancing} and talent evaluation~\cite{li2023ezinterviewer}.

\begin{figure}[t]
 \centering
  \includegraphics[width=\linewidth]{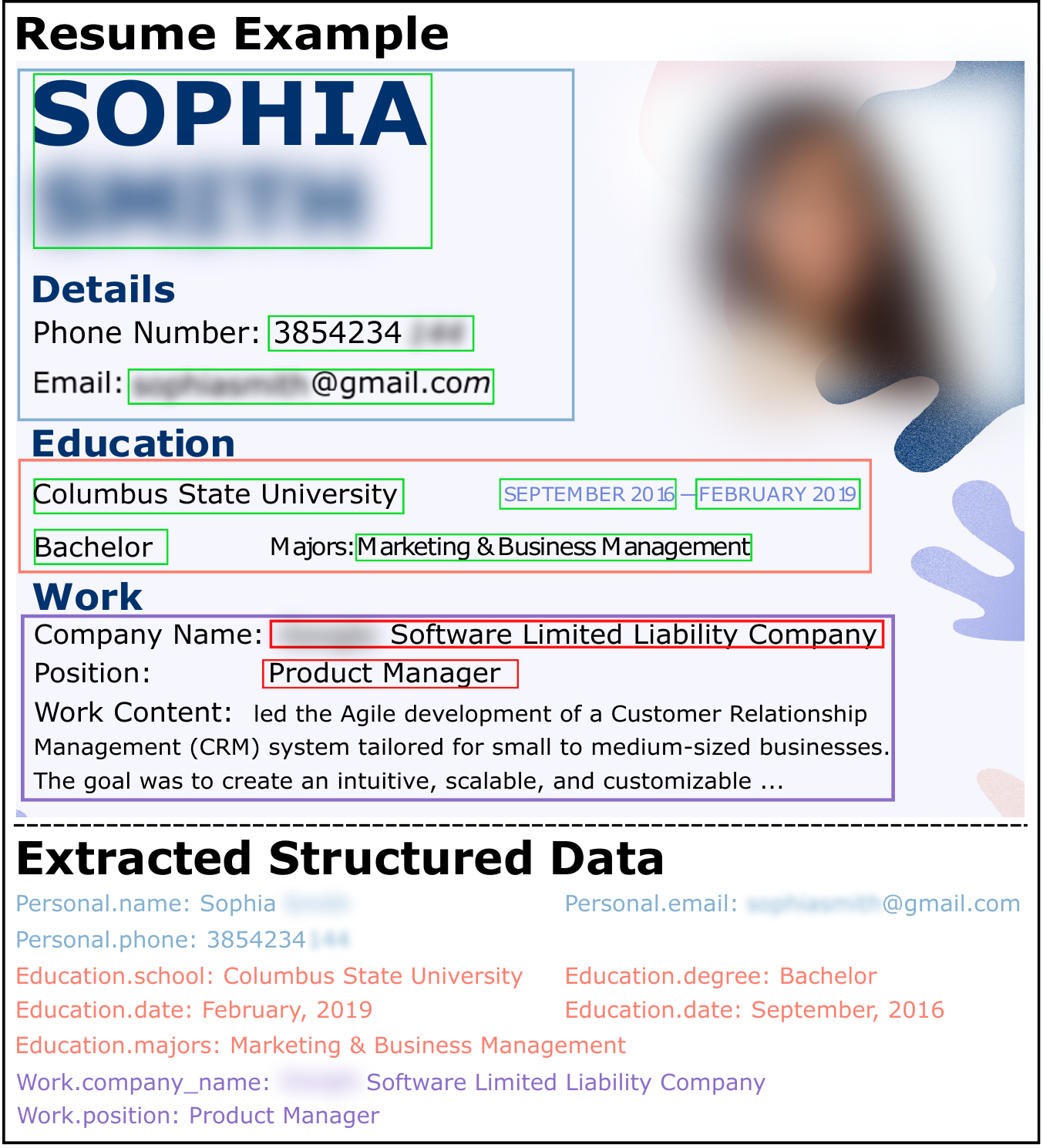}
  \caption{An illustrative example of resume understanding.}
  \label{fig:intro}
\end{figure}

As illustrated in Figure~\ref{fig:intro}, a resume conventionally incorporates the candidate's personal information (e.g., name), educational background (e.g., major, school), work experience, and various other pertinent details. The conversion of resumes into textual formats can be accomplished by utilizing optical character recognition~(OCR), thereby rendering resume understanding as a text-mining task~\cite{chandola2015online}. Traditional rule-based approaches extract structured information from resumes via various techniques, including keyword matching~\cite{kopparapu2010automatic} and lexical analysis~\cite{siefkes2005overview}. Furthermore, several studies regard it to be a sequence labeling task, wherein techniques such as hidden Markov models (HMM)~\cite{wentan2017chinese} and conditional random fields (CRF)~\cite{chen2016information} are employed to forecast the category of each word within a textual sequence, such as \textsl{personal.name} and \textsl{education.major}, as shown in Figure~\ref{fig:intro}. 

In recent times, the utilization of pre-trained models for document understanding has become widespread~\cite{xu2020layoutlm}. These models have attained state-of-the-art performance on a variety of benchmark tasks by integrating multi-modal features, such as visual and layout information.
Despite this, they still face several challenges in addressing the problem of resume understanding. To begin with, these transformer-based models are usually built on token-level granularity, rendering them difficult to apply when parsing long documents such as resumes. Furthermore, there exists a hierarchical relationship among the information that needs to be extracted from resumes, as exemplified by the personal information block depicted in Figure~\ref{fig:intro}, which includes information like name, email, and phone number. However, prior document understanding models have not been capable of effectively modeling such hierarchical relationships.

To address the above challenges, we propose an efficient resume understanding model, namely ERU.
Specifically, we first extract textual, visual, and layout inputs from each resume through an OCR tool and integrate them with a layout-aware multi-modal fusion transformer. Then, we pre-train the model on a large number of unlabeled resumes with three well-designed self-supervised
objectives, including the masked language model, visual position alignment, and masked segment prediction. After that, we fine-tune the model with a multi-granularity sequence labeling task to extract the structured information from resumes. Finally, extensive experiments on a real-world
dataset clearly demonstrates the effectiveness of our ERU.

\section{Related Works}
\textbf{Resume Understanding.}
Resume understanding aims to extract semantically structured information from resume documents, which can facilitate a wide range of intelligent recruitment applications, such as talent evaluation~\cite{li2023ezinterviewer} and person-job matching~\cite{qin2018enhancing}. Typically, resume understanding can be formulated as a text-mining task by using OCR tools to convert resumes into text format. Along this line, traditional studies predominantly rely on rule-based solutions like keyword searching~\cite{kopparapu2010automatic} and lexical analysis~\cite{siefkes2005overview}, or sequence labeling solutions, like HMM~\cite{wentan2017chinese}. However, these methods usually suffer from high costs stemming from the need for expert-crafted feature engineering. Recently, researchers have attempted to investigate the problem using neural network-based techniques. For instance, \emph{Chen et al.}~\cite{chen2016information} combined a bidirectional long short-term memory (BiLSTM) neural network with a CRF layer to parse resumes by leveraging Word2Vec features. Despite their success, the valuable visual and layout information in the resume document is often ignored, resulting in sub-optimal performance. \\
\textbf{Multi-modal Document Understanding.}
Recently, the utilization of pre-trained models by integrating
multi-modal features for document understanding has gained widespread adoption~\cite{xu2020layoutlm,wang2022lilt}. In this direction, LayoutLM is the first to jointly learn textual and 2-D layout information in a unified model, with visual information extracted by Faster R-CNN also integrated into the token embeddings~\cite{xu2020layoutlm}. To facilitate computational efficiency, LayoutLMv3 replaces the visual feature extraction module with a simple linear projection head. On such basis, DocFormer~\cite{appalaraju2021docformer}, LayoutXLM~\cite{xu2021layoutxlm}, and LiLT~\cite{wang2022lilt} further use diverse techniques to combine image, text, and layout, and also investigate additional pre-training tasks to enhance the learning of multi-modal representations. However, these studies fail to achieve effective resume understanding, primarily due to their inability to handle lengthy resume documents that are often multi-paged and to consider the hierarchical relationship among the fields that require parsing. 

\section{Preliminary}

In this study, we consider the multi-modal inputs of a resume to parse its structured information. Specifically, given a resume $R$, we leverage an OCR tool~\footnote{https://github.com/pymupdf/PyMuPDF} to construct a sequence of segments $ S~=~\{s_i\}^{|S|}_{i=1}$. Each segment $s_i$ contains: 1) textual input $text_i$ that includes a sequence of words, i.e., $text_i =\{c_j\}^{|s_i|}_{j=1}$; 2) layout inputs $b_i$ and $p_i$ that represent the bounding box $(b^0_i, b^1_i, b^2_i, b^3_i)$~\footnote{In the bounding box, $(b^0_i, b^1_i)$ represents the position of the upper left and $(b^2_i, b^3_i)$ represents the position of the lower right.} and page number of the segment $s_i$ respectively; and 3) visual input $v_i$ that denotes the visual information that cropped based on $b_i$. We regard resume understanding as a sequence labeling task, where the $C=\{e_i\}^{|C|}_{i=1}$ denotes all the possible fields, such as \textsl{personal.name} and \textsl{education.major} in Figure~\ref{fig:intro}, and $L= \{l_i\}^{|S|}_{i=1}$ denotes the corresponding label sequence of $S$. The formal definition of the problem is as follows:

Give a set of resume documents $\mathcal{R}$, where each $R \in \mathcal{R}$ contains a sequence of segments $S$, the target of resume understanding is to learn a model $M$, which can predict the corresponding label $l_i$ for each segment $s_i~\in~S$ to achieve parsing the structure information in $R$.


\section{METHOD}

\subsection{Overview}
\begin{figure*}[t]
 \centering
  \includegraphics[width=0.8\linewidth]{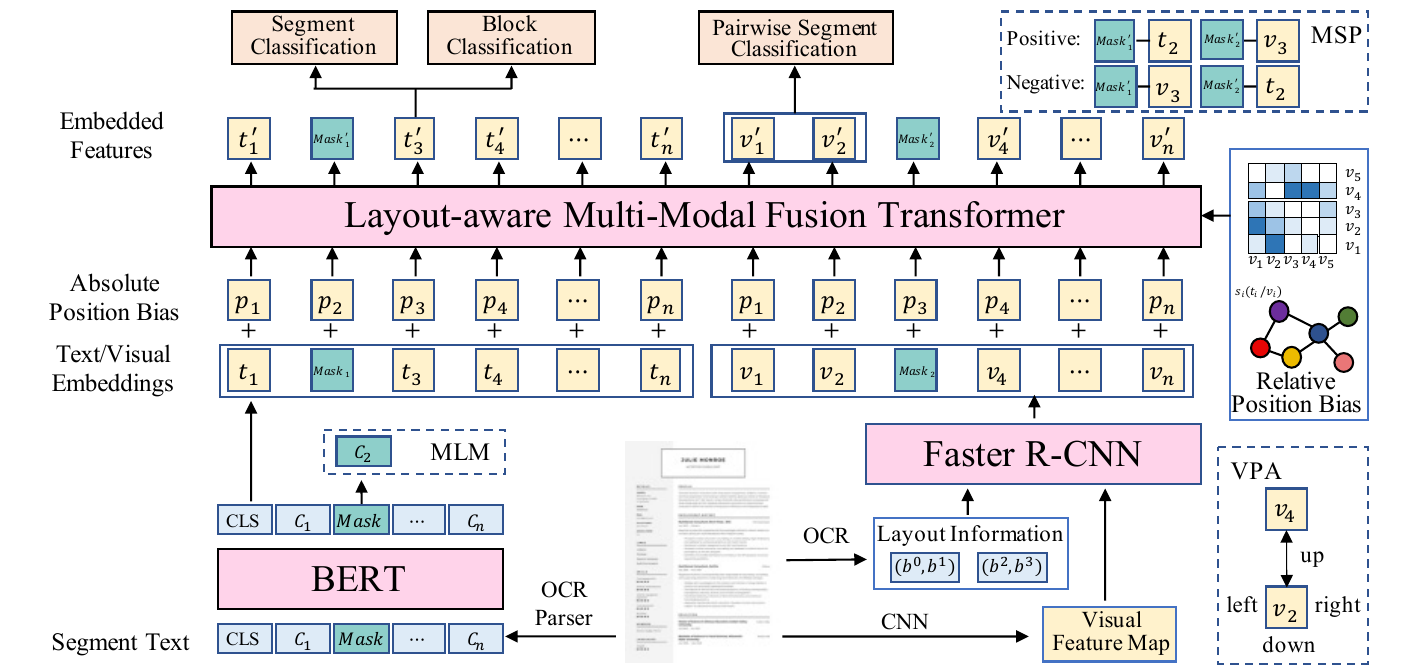}
  \caption{An illustration of the proposed ERU.}
  \label{fig:eru}
\end{figure*}

Figure~\ref{fig:eru} presents a comprehensive overview of our ERU.
To be specific, we first embed each segment from both textual and visual information. 
Then, a graph transformer is developed to model the interaction among different segments by representing each segment as two nodes, i.e., textual and visual nodes. 
Resume layout information is used to encode the spatial adjacency relationship among nodes.
Next, we introduce the pre-training strategy of our model, where three self-supervised objectives are defined.
Finally, the fine-tuning strategy on a relatively small-scale labeled dataset will be introduced based on the pre-trained model. 

\subsection{Multi-Modal Embedding}
We extract the textual and visual features from each segment.



\noindent \textbf{Textual Embedding.} 
To represent the text of one segment $s_i$, we used a 6-layer transformer which is initialized by BERT.
We add two specific tokens $[\text{CLS}]$ and $[\text{SEP}]$ to the beginning and end of the token sequence and use $[\text{CLS}]$ token as the text representation of the segment. As a result, the textual embedding vector can be derived by:
\begin{equation}
\small
\begin{aligned}
\mathrm{text_{i}} & = \left[[\mathrm{CLS}], c_1, c_2, \cdots, c_{|s_i|},[\mathrm{SEP}]\right],  \\ 
 {t_{i}}  & = \mathrm{BERT} \left( \mathrm{text_{i}} \right).
\end{aligned}
\end{equation}

\noindent \textbf{Visual Embedding.} 
To represent the visual information of each segment $s_i$, we enlarge the bounding box $b_i$ to $b_i^{\prime}$ by a certain proportion to capture the visual information $v_i$ such as the style and color of the text. We use Faster R-CNN~\cite{girshick2015fast} to extract visual features of the region where the segment is located, i.e., 
\begin{equation}
\small
v_{i}=\mathrm{Linear}_v(\mathrm{FasterRCNN}(I, b_i^{\prime})).
\end{equation}


\subsection{Layout-aware Multi-Modal Fusion Transformer}\label{sec:transformer}
To model the interaction among different segments and different modalities, we revisit each resume $S$ as a complete graph $G_s$ with the node set $X=\{t_0, v_0, ...,t_{|S|}, v_{|S|}\}$.
In other words, each segment  $s_i$ is donated as two nodes $\{t_i, v_i\}$ corresponding to the textual and visual features. Then, we develop a 4-layers graph transformer on the graph $G_s$ by inducing both the absolute position bias of each node and the relative position among nodes with the guidance of the resume layout information. 

\noindent\textbf{Absolute Position Bias.} For each segment $s_i$ ($t_i$ and $v_i$), there are three types of layout features, 1-D position, 2-D position, and the page number. The 1-D position is the absolute order $r_i$ obtained by parsing the resume from left to right and top to bottom. The 2-D position $b_i$ reflects the absolute position of the segment in the resume. We also encode the width and height of the segment since the same type of entities may have a similar size. We normalize the 2-D range to $(0, 1000)$. Thus the absolute position bias for each node $x_k = t_i$ or $v_i$ can be calculated by,
\begin{equation}
\small
\begin{split}
\mathrm{p}_k&=\mathrm{Pos_{2D}} \left(b_i, width, height, page\right) + \mathrm{Pos_{1D}(r_i)},\\
x_k^0 &= x_k + p_k,
\end{split}
\end{equation}
where $\mathrm{Pos_{2D}}$ and $\mathrm{Pos_{1D}}$ are implemented by MLP layers.
By adding the absolute position bias to node input, the self-attention mechanism in the transformer can catch the absolute layout information.

\noindent\textbf{Relative Position Bias.}
Indeed, the neighbor segments are often related, which indicates that the spatial adjacency relationship is also important to enhance the performance of resume understanding. Therefore, we follow Graphormer~\cite{ying2021transformers} and introduce the relative position bias to help the attention mechanism capture the neighbor information. Specifically, for each node pair $(x_m \in\{ t_i,v_i\}, x_n \in \{t_j,s_j\})$, the relative position bias $b_{\phi\left(m, n\right)}$ is defined with the distance between segment $s_i$ and $s_j$:
\begin{equation}
\small
\begin{aligned}
    b_{\phi\left(m, n\right)} = & W_x \left(|b^0_i - b^0_j|\right) + W_y \left(|b^1_i - b^1_j|\right), 
\end{aligned}
\end{equation}
where $W_x$, $W_y$ are learnable parameters. 

\noindent\textbf{Self-Attention.} Then, the self-attention mechanism can be defined as follows:
\begin{equation}
\small
\begin{aligned}
    \alpha_{m n} = &\frac{\left(x^0_m W_Q\right)\left(x^0_n W_K\right)^T}{\sqrt{d_f}}+b_{\phi\left(m, n\right)},\\
    x'_m = &\sum_n \frac{\exp \left(\alpha_{m n}\right)}{\sum_k \exp \left(\alpha_{m k}\right)}x^0_n   W_V,
\end{aligned}
\end{equation}
where $W_Q$, $W_K$, and $W_V$ are the learnable query, key, and value metrics in the transformer models.

\subsection{Pre-training Objectives} 
Given the extensive corpus of unlabeled resumes, we design a pre-training strategy that includes three self-supervised tasks, each with its specific objectives, as detailed below.
 
\noindent\textbf{Maked Language Model.}
The MLM task aims to make the text encoder adapt to the resume data. We initialize it with a pre-trained BERT model using 6 layers to leverage the existing knowledge.
We mask the selected tokens in one segment text but retain the corresponding layout features. The pre-training objective is to reconstruct the masked tokens which promotes the model to capture a contextual representation with the layout clues. The detailed computation method is similar to the word prediction task in BERT. We denote this loss as $L_{MLM}$.

\noindent\textbf{Visual-Position Alignment.}
To enhance the performance of extracting visual features, we propose to predict the relative position among segments with only their visual features, which encourages the visual features to capture the hidden relationship among neighbor segments.
For simplicity,  we divide the relative position into four directions $o = \{up, down, left, right \}$. For $v_i$, we randomly select a neighbor $v_{i}'$ and get a label $o_i$ according to their 2D position. We concatenate the two visual features and use a classification layer to predict the label. The $L_{V P A}$ is described as:
\begin{equation}
\small
\mathcal{L}_{V P A}=-\sum_{i=1}^{|S|} \log p_{\theta_{vpa}}(o_i \mid [v_i:v_i']),
\end{equation}
where $[:]$ denotes a concatenation operation, and the possibility function $p_{\theta}( b \mid a)$ is implemented by an MLP layer with parameter $\theta$ and the softmax function as the activation of the output layer, where $a$ is the input vector and the dimension of the output is same as the dimension of $b$. To simplify, similar symbols will not be explained again in the future.

\noindent\textbf{Masked Segment Prediction.} 
To enhance the model's ability to integrate different modalities and learn contextual information, we drew inspiration from the MLM method of randomly masking tokens and applied it to the input sequence of segments. This involved randomly masking segments and allowing the model to learn the representation of the masked segments through context.
Specifically, given the embedding set of textual and visual terms $X = \{ t_0,..., t_{|S|}, v_0, ..., v_{|S|} \}$, we randomly replace $q \left( q<|S| \right)$ term representations with a randomly initialized vector from both the textual and visual term sequence, ensuring that there is no overlap between the masked terms.
This allows for the use of corresponding visual information when the text is masked, and vice versa.
Thus we get a new input sequence with masked vectors $\hat{X} = \{ t_0, \hat{t_1},..., t_{|S|}, \hat{v_0}, v_1 , \hat{v_2},..., v_{|S|} \}$. 
After pulling $X$ and $\hat{X}$ into the transformer encoder defined in Section~\ref{sec:transformer}, we can get the output embedding set $X'$ and $\hat{X}'$, respectively. Then we use contrastive learning on the masked terms to minimize the distance between their representations in $X'$ and $\hat{X}'$. Specifically, for all masked term $x_m$, we follow\cite{oord2018representation} and use the following loss:
\begin{equation}
\small
\mathcal{L}_{M S P}=-\sum_{m} \log \frac{\exp \left(\operatorname{Sim}\left(x_{m}^{\prime}, \hat{x}'_{m}\right) / \tau\right)}{\sum_{k=1}^N \exp \left(\operatorname{Sim}\left(x_{m}^{\prime}, \hat{x}'_{k}\right) / \tau\right)},
\end{equation}
where $\operatorname{Sim(\cdot)}$ measures the similarity between two vectors, such as Cosine similarity,  $\hat{x}'_k$ is randomly selected from the set $\hat{X}'$ with the sampling number $N\ll |S|$, and $\tau$ is the temperature coefficient.

After the introduction of three self-supervised pre-training tasks, the final pre-training objective is defined as: $\mathcal{L}_{pre}= \lambda_{mlm}\mathcal{L}_{M L M}+  \lambda_{vpa}\mathcal{L}_{V P A}+  \lambda_{msp}\mathcal{L}_{M S P}. $ where $\lambda_*$ are hyper-parameters to balance the different pre-training tasks.

\subsection{Multi-Granularity Sequence Labeling}

After pre-training the model, we fine-tuned our model on a small-scale labeled dataset to classify the segments into corresponding labels.
Specifically, for the output term representation sequence $X' = \{ t_0^{\prime}, t_1^{\prime},..., t_{|S|}^{\prime}, v_0^{\prime}, v_1^{\prime}, v_2^{\prime},..., v_{|S|}^{\prime} \}$, we use the representations of textual terms to predict both the class $l^{seg}_i$ of the corresponding segment and its corresponding block $l^{block}_i$, which provide a  multi-granularity label prediction task.
Moreover, we also introduce another task to provide better performance by predicting whether the given segment pair $g'=(t'_m,t'_n)$ belongs to the same blocks. If so, the label $l_i^{pair}=1$, else 0. The final fine-tuning loss is as follows,
\begin{equation}
\small
\begin{aligned}
        \mathcal{L}_f=-\sum_{i=1}^{|S|} \left( 
\log p_{\theta_{c1}}\left(l_i^{seg} \mid t'_i  \right)
+ \log p_{\theta_{c2}}\left(l_i^{block} \mid t'_i \right)\right) \\
+ \sum_{g} (\log p_{\theta_{c3}}\left(l_{i}^{pair} \mid g' \right).
\end{aligned}\label{equ:mgsl}
\end{equation}

\section{Experiments}
\subsection{Datasets}
For the resume understanding experiments, we collected unlabeled resumes to pre-train our model. And we annotated the dataset for fine-tuning and test using an open-source PDF annotation tool named PAWLS\footnote{https://github.com/allenai/pawls}. 
The statistics of the dataset are summarized in Table ~\ref{tab:resume datasets}.

\subsection{Implementation}
We set the maximum text length of a single segment to 32 and the maximum number of segments in a resume to 256. We used AdamW as the optimizer with the weight decay set to 0.01. The learning rates for the pre-trained model and other linear layers were 5e-5 and 1e-3, respectively. The temperature coefficient $\tau$ in MSP is set to 2. We set $\lambda_{mlm}, \lambda_{vpa}$ and $\lambda_{msp}$ as 1,1 and 0.6 for the best performance. All experiments were conducted on  8 Tesla A800 80G GPUs. 


\begin{table}[t]
   
    \caption{Statistics of resume understanding datasets.}
    
    \label{tab:resume datasets}
    \centering
    \resizebox{\linewidth}{!}{
        \begin{tabular}{@{}ccccc}
        \toprule
         \multirow{2}{*}{Datasets} & Pre-training & \multicolumn{3}{c}{Fine-tune Documents} \\
         \cline{3-5}
         & Documents & Training  & Validation &  Test \\
        \midrule
         samples & 169,286  & 196       & 100     & 141\\
        avg  of seg number & 88.90  & 95.49 & 98.34 & 102.42\\
        avg of seg length & 18.94 & 16.49 & 17.22 &15.24\\
        avg of pages    & 1.95 & 2.31  & 2.15 & 2.41\\
        \bottomrule
        \end{tabular}
    }
\end{table}

\subsection{Baselines}
We totally selected 6 models as baselines.
$\textbf{BERT}$~\cite{li2021method} is a typical sequence labeling method. $\textbf{LiLT}$~\cite{wang2022lilt} is trained on a single language and then directly fine-tuned on other languages.
$\textbf{DocFormer}$~\cite{appalaraju2021docformer} is a multi-modal pre-training model that adds spatial relative features.
$\textbf{LayoutXLM}$~\cite{xu2021layoutxlm} is a multi-modal pre-training model for multilingual document understanding.
$\textbf{LayoutLMv3}$~\cite{huang2022layoutlmv3} is a general-purpose pre-training model for both text-centric and image-centric Document AI tasks with a unified architecture. 
 $\textbf{LLM IE}$~\cite{xie2023empirical} is a generative information extraction method utilizing ChatGPT's advanced capabilities. To protect data privacy, our implementaion employs the open-source LLM Baichuan-13b which is enhanced by a supervised fine-tuning(sft) strategy.

\begin{table}[t]
    \caption{Overall performance of resume understanding.} 
    \label{segment main results}
    \centering
    \resizebox{\linewidth}{!}{
    \begin{tabular}{c c c c }
    \toprule
        Model & Precision & Recall &  F1  \\ \midrule
        LLM IE   & 63.37 $\pm$ 1.37 &  59.67 $\pm$ 2.03   &  61.24 $\pm$ 1.94 \\ 
        BERT   & 74.98 $\pm$ 2.21 &  79.36 $\pm$ 1.82   &  77.11 $\pm$ 0.54 \\ 
        LiLT  &  75.98 $\pm$ 1.16  &  89.47 $\pm$ 0.60  &  82.03 $\pm$ 0.50  \\ 
        DocFormer&  83.81 $\pm$ 2.66  &  86.05 $\pm$ 4.50   &  83.45 $\pm$ 0.64  \\
        LayoutXLM & 78.93 $\pm$ 1.66  & 89.14 $\pm$ 1.92 & 83.62 $\pm$ 0.28 \\ 
        LayoutLMv3 &  82.96 $\pm$ 1.53  &  86.29 $\pm$ 2.66  &  84.36 $\pm$ 0.53 \\ 
        ERU  & \textbf{84.64 $\pm$ 0.32} & \textbf{91.08 $\pm$ 0.52} &  \textbf{87.75 $\pm$ 0.22}\\ 
        \bottomrule
    \end{tabular}
    }
\end{table}

\begin{table}[t]
    \caption{Ablation experiments of different modules.} 
    \label{ablation results}
    \centering
    \resizebox{\linewidth}{!}{
    \begin{tabular}{c c c c }
    \toprule
        Model & Precision & Recall &  F1  \\ \midrule
ERU  & \textbf{84.64 $\pm$ 0.32} & \textbf{91.08 $\pm$ 0.52} &  \textbf{87.75 $\pm$ 0.22}\\ 
-w/o VPA &83.59 $\pm$ 0.15   & 90.47 $\pm$ 0.15 & 86.89 $\pm$ 0.12  \\
-w/o MSP & 81.20 $\pm$ 0.11 & 89.35 $\pm$ 0.15  &  85.07 $\pm$ 0.17\\
-w/o MSP+VPA &  83.79 $\pm$ 0.11 & 85.53 $\pm$ 0.16  &  84.65 $\pm$ 0.11\\
- w/o visual-emb & 84.53 $\pm$ 0.46  &  89.66 $\pm$ 0.45 & 87.02 $\pm$ 0.14  \\
- w/o relative-pos &  84.15 $\pm$ 0.42 &90.16 $\pm$ 0.62  & 87.05 $\pm$ 0.18 \\
- w/o multi-gran&  84.41 $\pm$ 1.68 & 89.07 $\pm$ 0.26  & 87.20 $\pm$ 0.03 \\

        \bottomrule
    \end{tabular}
    }
\end{table}


\subsection{The Overall Performance}
To evaluate the performance of the resume understanding task, following\cite{chen2016information}, we utilize precision, recall, and F1-score as the evaluation metrics. The comparison results for ERU and baseline methods are shown in Table~\ref{segment main results}. According to the results, we observe that our model outperformed all the baselines, which demonstrates the effectiveness of ERU. In comparison to the top-performing LayoutLMv3, we find that ERU achieves above 1.68\%, 4.79\%, and 3.39\% improvement on the precision, recall, and F1-score respectively. LayoutLMv3 operates on a token-level granularity and will divide long resumes into parts for individual processing, hindering its ability to fully comprehend a resume. In contrast, ERU is designed on a segment-level granularity, enabling it to understand a resume in its entirety. Additionally, ERU captures the hierarchical structure within resumes through a multi-granularity sequence labeling task. As a result, our model outperforms all baseline models.


We have also conducted further experiments to compare generative information extraction with our discriminative information extraction approach.  As shown in Tabel~\ref{segment main results}, LLM based method still falls short of extractive models when dealing with complex and lengthy texts. In the experiments, we found that certain flaws in LLM, such as the hallucination phenomenon\cite{jiang2024rag,xu2023large}, lead to the prediction of information not pertaining to the current resume. This is detrimental to the high accuracy required for resume parsing. Additionally, LLM's parsing efficiency is notably lower compared to discriminative models, highlighting the need for further research into LLM IE. In future work, we aim to focus on improving both the efficiency and accuracy of information extraction methods based on LLM.


\subsection{Ablation Studies}
To validate the effectiveness of each component in ERU, we conducted a series of six ablation experiments. The first set of experiments involved comparing ERU with three of its variants which represent the removal of the corresponding pre-training tasks: ERU \textsl{\mbox{-w/o VPA}}, \textsl{\mbox{-w/o MSP}}, and \textsl{\mbox{-w/o MSP+VPA}}. The second set of experiments compared ERU with three other variants: 1) \textsl{\mbox{-w/o visual-emb}}, which omits the visual embedding in multi-modal embedding; 2) \textsl{\mbox{-w/o relative-pos}}, which excludes the relative position bias in the layout-aware multi-modal fusion transformer; and 3) \textsl{\mbox{-w/o multi-gran}}, which uses only the segment classification loss in multi-granularity sequence labeling.

The first three ablation studies, which focus on pre-training losses, provide significant insights, as detailed in Table~\ref{ablation results}. Initially, the exclusion of the VPA pre-training task led to a noticeable decrease in ERU's performance, highlighting the role of high-quality visual feature extraction in enhancing the model's classification effectiveness. Furthermore, the absence of the MSP pre-training task resulted in an even more pronounced performance decline, emphasizing the importance of the model's ability to integrate features from different modalities during pre-training. By utilizing both MSP and VPA pre-training tasks, our model achieved nearly a 2.6\% improvement in F1-score, underlining the effectiveness of these designed pre-training tasks.

And the remaining three ablation experiments also clearly demonstrate their contributions to the overall performance as shown in Table~\ref{ablation results}. The removal of the visual embedding (\textsl{\mbox{-w/o visual-emb}}) impacts the model's effectiveness in processing and integrating visual information. Likewise, excluding the relative position bias (\textsl{\mbox{-w/o relative-pos}}) affects the model's understanding of spatial relationships in the data. Lastly, employing only the segment classification loss in sequence labeling (\textsl{\mbox{-w/o multi-gran}}) restricts the model's capability to label sequences at different levels of granularity, which influences the overall performance.

\section{Conclusion}
In this paper we introduced a novel efficient resume understanding model, ERU, to automatically extract structured information from resume documents. To be more specific, we first designed a layout-aware multi-modal fusion transformer to encode the segments in the resume with text, visual and lauoyt features. Then we pre-trained the model on a large number of unlabeled resumes with three self-supervised tasks. After that, we fine-tuned the pre-trained model to extract structured information in resume with a multi-granularity sequence labeling task. Finally, extensive experiments have clearly demonstrated the effectiveness of ERU.

\bibliographystyle{IEEEbib}
\bibliography{sample-base}

\end{document}


\sloppy

\appendix
\section{Appendix}

\subsection{Complexity Analysis}
To verify the superior efficiency of our model compared to token-level document understanding approach, we analyze the time complexity of ERU. In ERU, the primary computation cost lies in the textual embedding, visual embedding and Layout-aware Multi-Modal Fusion Transformer.
The complexity of textual embedding part is:
\begin{equation}
    \mathcal{T}_{t} = O(L_1|S|Q^2),
\end{equation}
where $L_1$ is the layer number of BERT, $|S|$ is the number of segment in one document, $Q$ is the max token number in one segment. For the visual embedding part, we utilize CNN to extract the features. The visual model complexity is:
\begin{equation}
    \mathcal{T}_{v} = O(L_{2}|S|E^2K^2I_{in}I_{out})
\end{equation}
 where $L_2$ is the layer number of CNN network, $E$ represents the side length of the convolutional kernel output feature map, $K$ denotes the side length of each convolutional kernel, $I_{in}$ is the number of input channels, and $I_{out}$ is the number of output channels. As for the final multi-modal fusion transformer, it takes both text and visual features as input, resulting in a length of $2|S|$. Assuming the number of multi-modal fusion transformer layer is $L_2$, the overall complexity of the entire model is:
\begin{equation}
    \mathcal{T}_{ERU} = O(L_1|S|Q^2 + L_2|S|E^2K^2I_{in}I_{out} + L_3|S|^2)
\end{equation}

In order to make a fair comparison between models using different visual extractors and models using only text features, we consider the scenario where all models are built solely based on text features. Assuming an article has $N$ tokens, the time complexity for ERU using only text features is as follow:
\begin{equation}
    \mathcal{T}_{ERU}(N) = O(L_1NQ + L_3(\frac{N}{Q})^2)
\end{equation}

As for typical token-based resume understanding methods, they usually divide the long document into several parts for separate processing. Therefore, focusing solely on text features, their time complexity can be expressed as follow:
\begin{equation}
\begin{aligned}
    \mathcal{T}_{token}(N) =& O(L^{\prime}\frac{N}{Z}*Z^2) \\
        =& O(L^{\prime}ZN)
\end{aligned}
\end{equation}
where $Z$ is the max process length of token-based method, $L^{\prime}$ is the number of transformer layer.
In our setting, $L_1 < L^{\prime}$, $L_3 < L^{\prime}$, $Q \ll Z$ and $N$ is usually around 2,000. As a result, the obtained value of $\frac{\mathcal{T}_{ERU}(N)}{\mathcal{T}_{token}(N)}$ is less than 0.1, clearly demonstrating that our method exhibits significantly lower time complexity compared to the token-based approach.

\subsection{LLM IE}
We applied the sft method to Baichuan-13b model, utilizing identical training data. This approach enabled the model to effectively generate structured information from the input resume texts. We utilize a parameter-efficient fine-tuning approach called LoRA~\cite{hu2022lora}. This method entails fixing the parameters of the pre-trained LLM and training rank decomposition matrices specific to each layer in the Transformer architecture. The learning objective is formulated as follow:
\begin{equation}
 \max_{{\Theta^L}^{\prime}} \sum_{(x, y) \in \mathcal{T}} \sum_{t=1}^{|y|} \log \left(P_{{\Theta^L}+{\Theta^L}^{\prime}}\left(y_t \mid x, y_{<t}\right)\right),
\end{equation}
where  $\Theta^L$ is the original parameters of LLM, ${\Theta^L}^{\prime}$ is the LoRA parameters which will be updated during the training process, $x$ and $y$ represent the ``Instruction Input" and ``Instruction Output" in the training set $\mathcal{T}$, $y_t$ is the $t-th$ token of $y$, $y_{<t}$ represents the tokens before $y_t$.

And the designed instruction is as follow:
\begin{tcolorbox}[colback=white,colframe=black!50!white,arc=2mm,boxrule=1pt,boxsep=0pt,left=6pt,right=6pt,top=6pt,bottom=6pt,boxrule=1pt,title={Instruction for resume understanding sft},fonttitle=\bfseries]
\texttt{\small Task Description: You will be provided with a job applicant's resume text. Your task is to extract structured information from the resume according to the following categories, excluding the category "Other". \\
Category Definitions: \\
- Other: Text that does not belong to any of the categories below.\\
- Name: The job applicant's name.\\
- Company Name: Name of the workplace or company.\\
...\\
Output Format: For each entity in the resume, classify it in the "Text:Category" format.\\
Now, please categorize the following resume content:}
\end{tcolorbox}

\subsection{Parameter Analysis}
For $\lambda_{mlm}$, $\lambda_{vpa}$ and $\lambda_{msp}$, we conducted coefficient experiments by setting $\lambda_{mlm}$ and $\lambda_{vpa}$ to 1 and varying $\lambda_{msp}$. Table \ref{para results} illustrates the results, which show that the model performs best when the parameter is set to 0.6, and the model's fluctuations are also minimal.

\begin{table}[h]
    \caption{The impact of the coefficient $\lambda_{msp}$.} 
    \label{para results}
    \centering
    \begin{tabular}{ccccccc}
    \toprule
             & 0.2 & 0.4 &  0.6 & 0.8 & 1.0 & 1.2 \\ \midrule
        F1   & 86.91 & 87.28 & \textbf{87.75} & 87.07 & 87.03 & 87.06 \\ 
        \bottomrule
    \end{tabular}
\end{table}

\subsection{Adjacent Features of Segments}

\begin{figure}[h]
  \vspace{-5mm}
  \includegraphics[width=\linewidth]{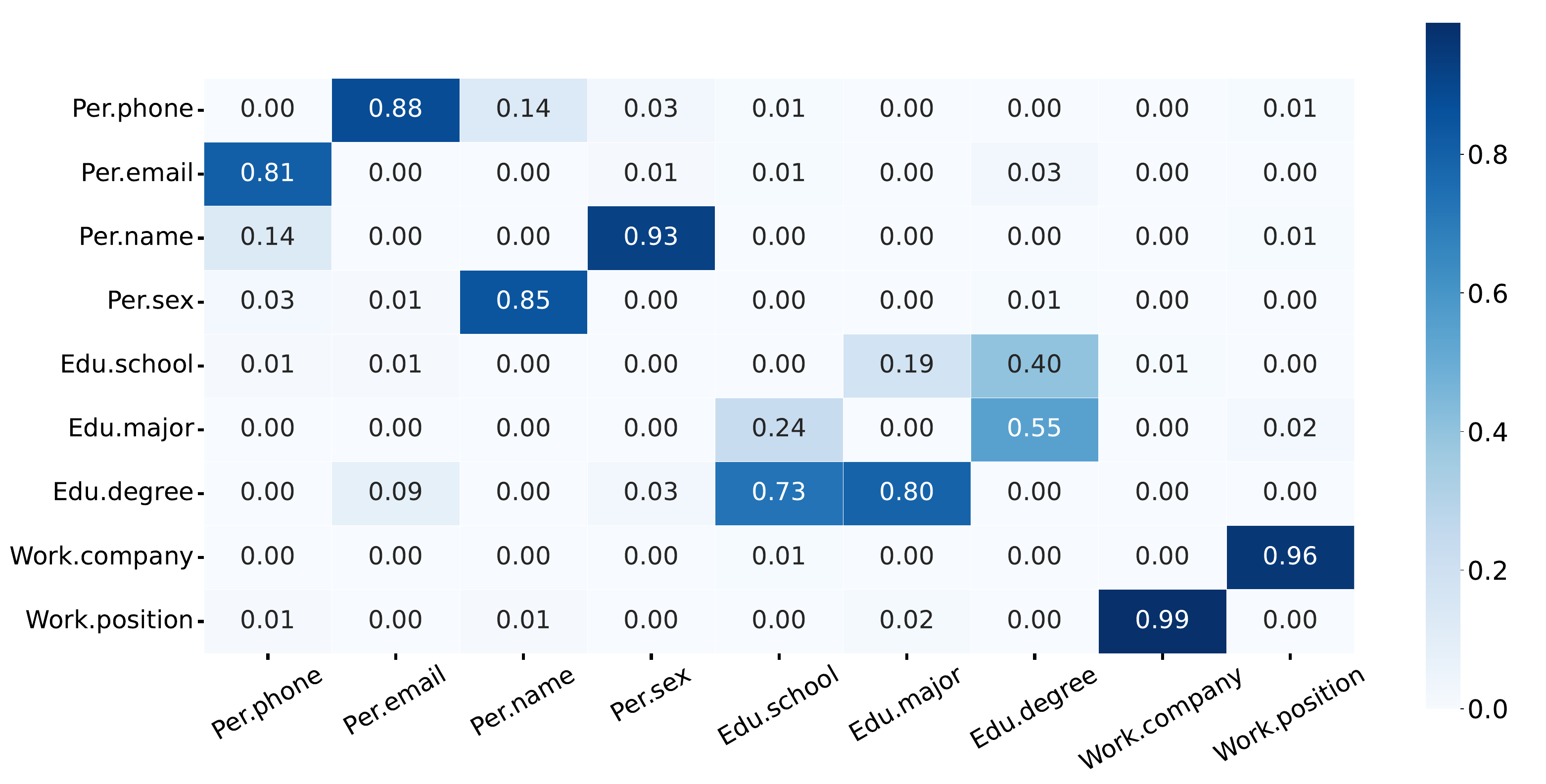}
  \vspace{-6mm}
  \caption{Heatmap showing the nearest neighboring segments for each kind of segment.}
  \label{fig:heat_maps}
\end{figure}

Figure \ref{fig:heat_maps} presents the nearest neighboring segments distribution. We can observe that some types of segments are often related and they exhibit adjacent features in terms of positional distance. Therefore, we follow Graphormer~\cite{ying2021transformers} and introduce the relative position bias to help the attention mechanism capture the neighbor information.


\vspace{-4mm}
\subsection{Case Study}
\begin{figure}[h]
  \includegraphics[width=\linewidth]{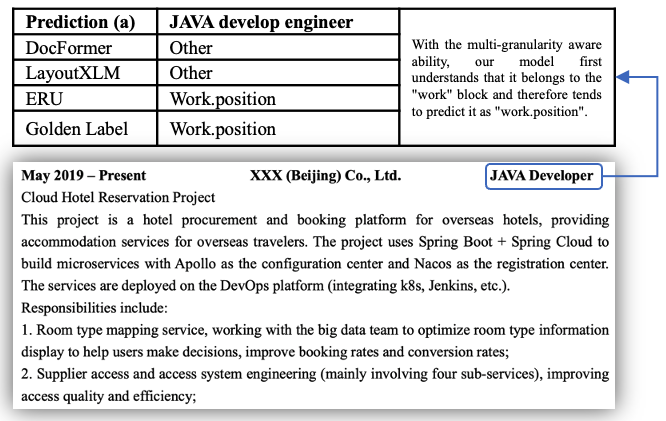}
  \caption{The case study of resume understanding via ERU and the baselines.}
  \label{fig:case}
\end{figure}

To further illustrate the effectiveness of our approach, we show an example of the test set in the aforementioned resume understanding experiments. The results presented in Figure \ref{fig:case} are obtained from the final models, which include ERU and the baselines. We can observe that ERU correctly recognized ``JAVA develop engineer'' as ``Work.position'', whereas the baselines predicted it as ``Other''. We also found that our model identified this segment as belonging to the ``Work'' block, which validates that the proposed multi-granularity sequence labeling can improve the effectiveness of resume understanding. 

\bibliographystyle{IEEEbib}
 \vspace{-0.3cm}
\bibliography{sample-base}